\documentclass[12pt,a4paper]{article}
\usepackage{graphicx}

\newtheorem{proposition}{Proposition}
\title{\bf Heuristic solutions to robust variants of the minimum-cost integer flow problem}
\author{Marko \v Spoljarec\\
       {\normalsize {\em Privredna banka Zagreb}}\\
       {\normalsize {\em Radni\v cka cesta 44, 10000 Zagreb, Croatia}}\\
       {\normalsize {\em E-mail: marko.spoljarec@pbz.hr}} \\
       \\
       Robert Manger \\
       {\normalsize {\em University of Zagreb, Faculty of Science, Department of Mathematics}}\\
       {\normalsize {\em Bijeni\v{c}ka cesta 30, 10000 Zagreb, Croatia}}\\
       {\normalsize {\em E-mail: manger@math.hr}}}
\date{ }

\begin{document}

%
%        Title and abstract
%

\maketitle

\begin{quote}
  {\bf Abstract.} This paper deals with robust optimization applied to network
  flows. Two robust variants of the minimum-cost integer flow problem are considered.
  Thereby, uncertainty in problem formulation is limited to arc unit costs and
  expressed by a finite set of explicitly given scenarios. It is shown that both
  problem variants are NP-hard. To solve the considered variants, several heuristics
  based on local search or evolutionary computing are proposed. The heuristics are
  experimentally evaluated on appropriate problem instances.
  \vspace{2ex} 
\end{quote}

%
%        Section 1
%

\section{Introduction}

{\em Flows in networks\/} are an important modeling paradigm used in optimization.
Models based on flows are applied in different areas, such as transport, logistics,
production planning, network design, etc. There are several types of network flow
problems, but most of them reduce to the relatively general {\em minimum-cost flow
problem\/}.

As it is true for any optimization task, an instance of the minimum-cost flow problem
is specified by exact values of parameters within its objective function and its
constraints. However, in real-world situations parameter values are often hard to
determine since they may depend on unpredictable future circumstances or perhaps cannot
be measured accurately. Then we speak about {\em uncertainty\/} in problem formulation.
The traditional approach to optimization tends to ignore uncertainty. It means that
parameters are very often given approximate, unreliable or ad-hoc values. Unfortunately,
such approach can lead to inferior or even infeasible solutions. Therefore, instead of
ignoring uncertainty, it is much better to admit its existence and find a more appropriate
way of dealing with it.

A state-of-the-art approach to deal with the mentioned uncertainty is called {\em robust
optimization\/}. According to that approach, a finite or infinite set of {\em scenarios\/} is
defined - each of them specifies a possible combination of parameter values. Only those solutions
are considered that are feasible for all scenarios. The behavior of any solution under any scenario
is measured in some way. For each solution, its worst behavior over the whole set of scenarios is
recorded. As the optimal solution in the robust sense, the one is chosen whose recorded
worst behavior happens to be the best among all solutions. A consequence of using the robust
approach is that the initial (let us say) minimization problem transforms into a more complex
min-max problem. The robust solution does not need to be really optimal for any scenario,
but it is chosen in order to be acceptable even in the most adverse circumstances.

The foundations of robust optimization have been laid out in the seminal works
\cite{Ben-Tal,Bertsim1,Bertsim2,Kouvelis}. For our purposes the most important reference
is the book \cite{Kouvelis}, which provides a framework for robust {\em discrete\/} (or
{\em combinatorial\/}) optimization. More recent surveys of the whole discipline and
its general results are available in \cite{Aissi1, Aissi2,Bertsim3,Kasperski}.

The existing research papers on robust discrete optimization are mostly concerned
with the shortest path problem. Also well covered is the knapsack or the minimum spanning
tree problem. There are not too many works dealing with flows in networks. The available
papers on robust flows
\cite{Aissi3,Atamturk,Bertsim1,Bertsim4,Boginski,Minoux1,Minoux2,Ordonez,Poss,Righetto,Rui}
are hard to compare since they use quite different definitions and concepts. Some of them,
e.g.\ \cite{Boginski, Rui}, capture uncertainty in parameters by assuming a
finite and explicitly given set of scenarios. Other papers, e.g.\
\cite{Aissi3,Atamturk,Bertsim1,Minoux1,Minoux2,Ordonez,Poss},
assume intervals or even more general geometric constructs for parameter values, thus
dealing implicitly with an infinite but rather regular set of scenarios.
There are also big differences regarding the scope or extent of
uncertainty.

It is well known that the conventional (non-robust) flow problems can be solved
efficiently by polynomial-time algorithms. On the other hand, many robust variants
of flow problems turn out to be NP-hard. Consequently, finding efficient algorithms
for robust variants is a challenging and important research topic. At this moment,
there are not many algorithms found in literature that can be regarded as efficient
and suitable for real-world situations. Authors of the available papers have mainly
been concerned with complexity issues or specialized solutions. For instance, the
authors of \cite{Rui} propose an algorithm for solving a robust minimum-cost integer
flow problem, which relies on an unspecified sub-algorithm for solving the corresponding
(NP-hard) robust shortest path problem. As far as we know, there are no reports in
literature on solving robust minimum-cost flow problems by standard heuristics or
meta-heuristics.

The aim of this paper is to demonstrate that robust flow problems can be solved with
reasonable efficiency even if they are NP-hard. The aim is also to present useful
and practically relevant algorithms for solving some types of robust flow problems.
In order to fulfill its aims, the paper considers two robust variants of the
minimum-cost {\em integer\/} flow problem. Both of them are based on an explicitly
given finite set of scenarios. Uncertainty expressed through scenarios is limited
to arc unit costs. It is shown that both problem variants are NP-hard, which is
an indication that they could be solved efficiently only by approximate algorithms.
The paper proposes several heuristic solutions based on local search and on
evolutionary computing, respectively. The proposed heuristics are tested on
carefully constructed problem instances, and their performance is measured
in terms of accuracy and speed.

Our decision to study heuristics instead of some other types of algorithms is
motivated by the fact that real-world problem instances may be fairly large.
It is true that smaller instances can be solved exactly by general-purpose
optimization software packages. Also, moderately sized tasks can be solved
at least approximately by relaxation (ignoring integrality constraints)
and rounding. Still, very large instances may easily become intractable
for general-purpose methods even in the relaxed form, so that they could
be tackled only by heuristics.

Apart from this introduction, the rest of the paper is organized as follows.
Section~2 specifies two robust variants of the minimum-cost integer flow problem
and shows that both of them are NP-hard. Section~3 describes some basic procedures
with flows that will further on be used as building blocks for heuristics.
Section~4 presents our heuristics for solving the considered problem variants.
Section~5 reports on experimental evaluation of the heuristics. The final
Section~6 gives conclusions.

%
%        Section 2
%

\section{Problem variants and their complexity}

We first describe the {\em conventional\/} (non-robust) variant of the minimum-cost integer
flow problem. We use the formulation from \cite{Papa}. Let $G=(V,A)$ be a {\em network\/}
(directed graph), where $V=\{v_1,v_2, \ldots, v_n\}$ is a set of $n$ elements
called {\em vertices\/} and $A\subset V\times V$ is a set of ordered pairs of vertices
called {\em arcs\/}. Each arc $(v_i,v_j)$ is characterized by its {\em capacity\/} $u_{ij}$
and its {\em unit cost\/} $c_{ij}$. Vertex $v_1$ is called the {\em source\/} and vertex
$v_n$ the {\em sink\/}. We consider {\em feasible flows\/} that transfer a required
amount of flow $F$ from the source to the sink. Thereby, a feasible flow is a nonnegative
function defined on arcs, which conforms to the {\em flow conservation rule\/} in vertices and to
the {\em capacity constraints\/} along arcs. The arc flow value assigned to an arc $(v_i,v_j)$
is denoted by $x_{ij}$. The flow cost is obtained by summing the products $c_{ij}x_{ij}$
over all arcs $(v_i,v_j)$. The objective is to find a feasible flow with {\em minimal
cost\/}.

In this paper we assume that all input data $u_{ij}, c_{ij}, F$ are nonnegative
integers. Moreover, we expect that the flow itself should consist of integer arc
values $x_{ij}$. Thus we indeed consider the minimum-cost {\em integer\/} flow
problem. The restriction to integers makes sense when a discrete phenomenon is
modeled, such as transportation of packaged goods, assignment of tasks to agents,
etc.

The described minimum-cost integer flow problem can more formally be defined as
the following {\em integer linear programming\/} problem \cite{Papa}:
\begin{eqnarray*}
  % & & \\
  \mbox{MCIF} &\;\ldots\;& z =\sum_{(v_i,v_j)\in A}c_{ij}x_{ij} \longrightarrow \min \\
  & & \\
  & & \mbox{subject to:} \\
  & & \sum_{v_j \in V \atop (v_i,v_j)\in A}x_{ij} -
      \sum_{v_j \in V \atop (v_j,v_i)\in A}x_{ji} = \left\{\begin{array}{l}
                                                              F \mbox{ if } v_i=v_1\\
                                                              -F \mbox{ if } v_i=v_n\\
                                                              0 \mbox{ otherwise }
                                                             \end{array} \right.,
                                                             \mbox{ for all } v_i \in V \\
  & & 0\leq x_{ij} \leq u_{ij} , \mbox{ for all } (v_i,v_j)\in A \\
  & & x_{ij} \mbox{ integer}, \mbox{ for all } (v_i,v_j)\in A \\
\end{eqnarray*}

Note that the considered minimum-cost flow problem includes also the well-known
minimum-cost maximal flow problem \cite{Carre} as a special case. Indeed, for a
given network we can compute its maximal flow value in advance. Then, in order to
solve the minimum-cost maximal flow problem, we can solve our problem with $F$
set to the computed maximal flow value. Note also that in some literature, e.g.\
\cite{Bazaraa,Jung,Korte}, the minimum-cost flow problem is described in a
different way allowing more sources and sinks. Although seemingly more general,
such alternative formulation can easily be reduced to ours. Reduction is obtained
by introducing an additional source and an additional sink, and by adding arcs with
appropriate capacities from the new source to each of the original sources, as well
as from each of the original sinks to the new sink.

Now we describe two {\em robust\/} variants of the minimum-cost integer flow problem.
According to our adopted approach from \cite{Kouvelis}, uncertainty in input data is
captured by a finite set of {\em scenarios\/} $S$. A particular scenario $s\in S$ is
expressed through a specific set of arc unit costs $c_{ij}^s$. We assume that network
structure and arc capacities are the same for all scenarios. Consequently, the set
of feasible flows also remains the same, no matter which scenario has been chosen.

The first robust variant is called {\em absolute\/} \cite{Kouvelis} or {\em min-max\/}
\cite{Aissi1} robust variant. There, the behavior of a feasible flow under a certain
scenario is measured absolutely, i.e.\ as the actual flow cost. For each feasible flow,
its worst behavior (i.e.\ its maximal flow cost) over all scenarios is recorded. As the
robust solution, the flow is chosen whose worst behavior is the best (i.e.\ minimal)
among all feasible flows. More formally, the absolute robust variant is defined as
follows:
\begin{eqnarray*}
  % & & \\
  \mbox{RMCIF-A} &\;\ldots\;&
  z =\max_{s\in S}\left\{\sum_{(v_i,v_j)\in A}c_{ij}^s x_{ij}\right\}\longrightarrow \min \\
  & & \\
  & & \mbox{subject to:} \\
  & & \sum_{v_j \in V \atop (v_i,v_j)\in A}x_{ij} -
      \sum_{v_j \in V \atop (v_j,v_i)\in A}x_{ji} = \left\{\begin{array}{l}
                                                              F \mbox{ if } v_i=v_1\\
                                                              -F \mbox{ if } v_i=v_n\\
                                                              0 \mbox{ otherwise }
                                                             \end{array} \right.,
                                                             \mbox{ for all } v_i \in V \\
  & & 0\leq x_{ij} \leq u_{ij} , \mbox{ for all } (v_i,v_j)\in A \\
  & & x_{ij} \mbox{ integer}, \mbox{ for all } (v_i,v_j)\in A \\
\end{eqnarray*}

The second robust variant is called robust {\em deviation\/} \cite{Kouvelis} or robust
{\em min-max regret\/} \cite{Aissi1} variant. There, the behavior of a feasible flow
under a certain scenario $s$ is measured as deviation of the actual cost from the
optimal cost $z^s$ for that scenario (computed in advance). Again, the flow is chosen
whose worst behavior over the whole set of scenarios is the best possible. Or more
formally, the robust deviation variant is defined in the following way:
\begin{eqnarray*}
  \mbox{RMCIF-D} &\;\ldots\;&
  z =\max_{s\in S}\left\{\sum_{(v_i,v_j)\in A}c_{ij}^s x_{ij} - z^s\right\}\longrightarrow \min \\
  & & \\
  & & \mbox{subject to:} \\
  & & \sum_{v_j \in V \atop (v_i,v_j)\in A}x_{ij} -
      \sum_{v_j \in V \atop (v_j,v_i)\in A}x_{ji} = \left\{\begin{array}{l}
                                                              F \mbox{ if } v_i=v_1\\
                                                              -F \mbox{ if } v_i=v_n\\
                                                              0 \mbox{ otherwise }
                                                             \end{array} \right.,
                                                             \mbox{ for all } v_i \in V \\
  & & 0\leq x_{ij} \leq u_{ij} , \mbox{ for all } (v_i,v_j)\in A \\
  & & x_{ij} \mbox{ integer}, \mbox{ for all } (v_i,v_j)\in A \\
\end{eqnarray*}

On the first sight, RMCIF-A and RMCIF-D seem to be nonlinear since their objective
functions involve min-max combinations. However, both problems can easily be
transformed into integer linear programming problems. Transformation is done by
introducing an additional variable $y$ in a manner shown in \cite{Kouvelis}. Indeed,
here are the corresponding ``linearized" versions RMCIF-A$^\prime$ and RMCIF-D$^\prime$,
respectively:
\begin{eqnarray*}
  \mbox{RMCIF-A}^\prime &\;\ldots\;& y \longrightarrow \min \\
  & & \\
  & & \mbox{subject to:} \\
  & & \sum_{(v_i,v_j)\in A}c_{ij}^s x_{ij} \leq y , \mbox{ for all } s\in S \\
  & & \sum_{v_j \in V \atop (v_i,v_j)\in A}x_{ij} -
      \sum_{v_j \in V \atop (v_j,v_i)\in A}x_{ji} = \left\{\begin{array}{l}
                                                              F \mbox{ if } v_i=v_1\\
                                                              -F \mbox{ if } v_i=v_n\\
                                                              0 \mbox{ otherwise }
                                                             \end{array} \right.,
                                                             \mbox{ for all } v_i \in V \\
  & & 0\leq x_{ij} \leq u_{ij} , \mbox{ for all } (v_i,v_j)\in A \\
  & & x_{ij} \mbox{ integer}, \mbox{ for all } (v_i,v_j)\in A
\end{eqnarray*}
\begin{eqnarray*}
  \mbox{RMCIF-D}^\prime &\;\ldots\;& y \longrightarrow \min \\
  & & \\
  & & \mbox{subject to:} \\
  & & \sum_{(v_i,v_j)\in A}c_{ij}^s x_{ij} \leq y + z^s, \mbox{ for all } s\in S \\
  & & \sum_{v_j \in V \atop (v_i,v_j)\in A}x_{ij} -
      \sum_{v_j \in V \atop (v_j,v_i)\in A}x_{ji} = \left\{\begin{array}{l}
                                                              F \mbox{ if } v_i=v_1\\
                                                              -F \mbox{ if } v_i=v_n\\
                                                              0 \mbox{ otherwise }
                                                             \end{array} \right.,
                                                             \mbox{ for all } v_i \in V \\
  & & 0\leq x_{ij} \leq u_{ij} , \mbox{ for all } (v_i,v_j)\in A \\
  & & x_{ij} \mbox{ integer}, \mbox{ for all } (v_i,v_j)\in A \\
  % & &
\end{eqnarray*}

Note that our two robust variants RMCIF-A and RMCIF-D differ only in the way how
the robust objective function is defined. In some contexts we will treat both
variants together by referring simply to the ``robust minimum-cost integer flow
problem". We will also use the common acronym RMCIF.

To illustrate the introduced problem variants, we give now a simple example.
Let us consider the network shown in Figure~1. All arc capacities are set to 1.
Vertex $v_1$ is the source and vertex $v_{14}$ the sink. The desired flow value is
$F=2$. Arc unit costs are given by two alternative scenarios and shown as arc
labels separated by ``/". We would like to compute the solution of the conventional
variant MCIF for each scenario. Also, we would like to solve the absolute robust
variant RMCIF-A, as well as the robust deviation variant RMCIF-D.

\begin{figure}[hbt]
\begin{center}
    \ \\
    \ \\
    \leavevmode
    \includegraphics[width=400pt]{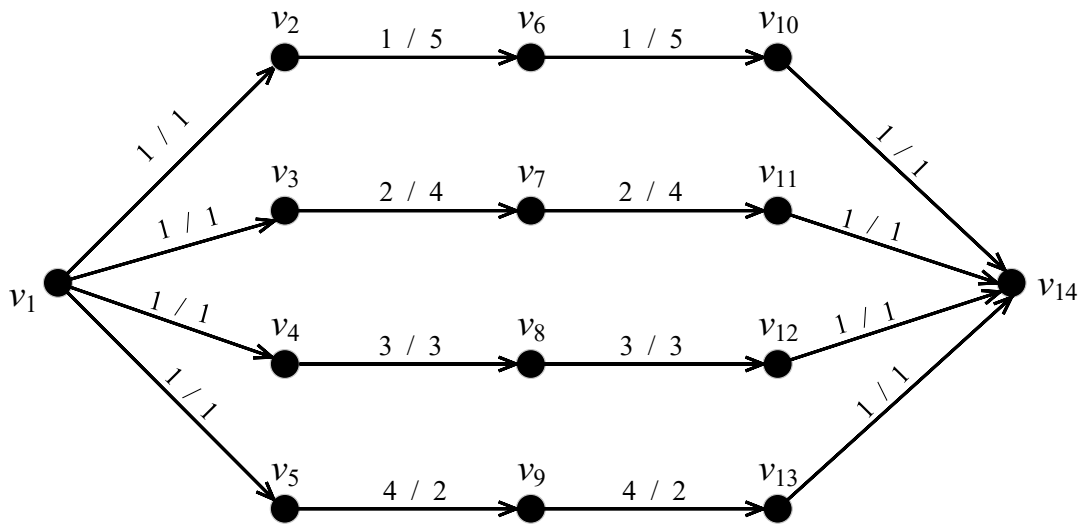}
    \caption{A sample problem instance with two scenarios.}
\end{center}
\end{figure}

As we can see, our network in Figure~1 consists of four separate paths. Each
path connects the source to the sink and has capacity 1. Thus any feasible flow
with value $F=2$ must be a combination of exactly two unit flows, each of them
sending one unit of flow through a distinct path.

Let flow~1 be the flow through the uppermost path in Figure~1, flow~2
through the next uppermost path, \ldots , flow~4 through the lowermost path.
By examining all six combinations of two among four flows, we can easily
check that the optimal solution under the first scenario is the combination
of flow~1 with flow~2 - the respective cost is 10. Similarly, the optimal
solution under the second scenario is flow~3 plus flow~4 with the cost l4.
On the other hand, the combination of flow~2 with flow~4 gives the optimal
solution in the sense of absolute robustness and its cost is 16. Finally,
the optimum in the sense of robust deviation can be achieved either by
flow~1 plus flow~4, or by flow~2 plus flow~3 - the optimal deviation
of cost is 4.

The presented example clearly shows that a robust solution can differ
from any conventional solution corresponding to a particular scenario.
Also, we see that the considered two criteria of robustness can produce
different results.

Now we will discuss {\em computational complexity\/} issues \cite{Garey,Papa}
regarding different variants of the minimum-cost integer flow problem. It is
well-known \cite{Jung} that the conventional (non-robust) variant MCIF can be
solved in polynomial time, even if integral solutions are required. However,
by switching to robust versions, the problem becomes much more complex, as
stated by the following proposition.

\begin{proposition}
  The problem variants RMCIF-A and RMCIF-D are both NP-hard. This claim
  is true even if the number of scenarios is limited to 2.
\end{proposition}
The proof is obtained by polynomial reduction \cite{Garey,Papa} of suitable
shortest path problem variants to our considered flow problem variants. Namely,
the standard source-to-sink shortest path problem is equivalent to the minimum-cost
integer flow problem posed in the same network, with arc lengths interpreted
as unit costs, and with the required flow value $F=1$. Indeed, for $F=1$,
due to integrality, each flow must be a unit flow. Any unit flow determines
a path from the source to the sink, and vice-versa. Thus the minimum-cost
unit flow is equivalent to the shortest path. Moreover, the solution of
any robust variant of the minimum-cost unit flow problem is equivalent
to the solution of the corresponding robust variant of the shortest path
problem. On the other hand, it has been proved in \cite{Kouvelis} that
both robust variants of the shortest path problem are NP-hard, even in
the case with only 2 scenarios. Therefore the corresponding network flow
variants must also be NP-hard.

In spite of the observed NP-hardness, smaller instances of RMCIF-A or
\mbox{RMCIF-D} can be solved exactly by general-purpose integer programming
software packages. There\-by, ``linearized" formulations RMCIF-A$^\prime$
and RMCIF-D$^\prime$, respectively, should be used. Still, larger problem
instances can easily become intractable. The only practical way of
dealing with such larger instances is switching to approximate solutions.

%
%        Section 3
%

\section{Basic procedures with flows}

In this section we present some basic computational procedures dealing
with flows. Most of them are relatively well known since they are
incorporated within standard networking algorithms. However, our intention
is to use them in a novel way as building blocks for heuristics. The
procedures are listed and specified in Table~1. It is assumed that all
involved flows operate on the same network.

\begin{table}[p]
\begin{center}
\begin{tabular}{|l||l|l|}
  \hline
  Procedure   & Input  & Output   \\
  name        &        &          \\
  \hline\hline
  Summation & A list of flows & The pseudo-flow $g$ equal \\
  of flows  &   $f_1, f_2, \ldots , f_r$ & to the sum of $f_1, f_2, \ldots , f_r$ \\
  \hline
  Flow & An integer flow $f$ & A list  $\phi_1, \phi_2, \ldots , \phi_F$ of $F$ unit \\
  decomposition & with value $F$ & flows whose sum is equal to $f$ \\
  \hline
  Flow      & A list of flows & The (possibly non-integer) flow \\
  centering & $f_1, f_2, \ldots , f_r$ & $g$ equal to the arithmetic mean \\
            & & of $f_1, f_2, \ldots , f_r$ \\
  \hline
  Flow         & An integer flow $f$ & An integer flow $g$ with value \\
  augmentation & with value $F$ & $>F$, which can be considered \\
               &                & as an augmented version of $f$ \\
  \hline
  Flow     & A non-integer flow & An integer flow $g$, which can \\
  rounding & $f$ with value $F$ & be regarded as an integral \\
           & & approximation of $f$, and \\
           & & whose value is $F$ rounded \\
           & & to the nearest integer \\
  \hline
  Flow        & Two lists of unit & An integer flow $h$, which is a \\
  composition & flows, $\phi_1,\phi_2,\ldots,\phi_F$ & sum  of roughly equal number \\
              & and $\psi_1, \psi_2, \ldots , \psi_F$ & of unit flows from both input \\
              & & lists, and whose value is $F$ \\
  \hline
  Flow cost & A scenario $s$, an & An integer flow $g$, which can be \\
  reduction & integer flow $f$ with & considered as a cost-reduced  \\
            & value $F$ whose cost & version of $f$, having the same \\
            & under scenario $s$ & value $F$ but cost (under \\
            & is $C$ & scenario $s$) smaller than $C$ \\
  \hline
  Flow         & An integer flow $f$ & An integer flow $g$ with the same \\
  perturbation & with value $F$ & value $F$, which can be regarded \\
               &                & as a slightly perturbed version \\
               & & of $f$ \\
  \hline
  Flow          & Two integer flows & An integer flow $h$ with the same \\
  harmonization & $f$ and $g$, both & value $F$, which can be regarded \\
                & with value $F$ & as an adjusted version of $f$ \\
                & & being more ``similar" to $g$ \\
                & & than the original version \\
  \hline
  Finding a   & An integer $F$ & An arbitrary integer flow $f$ \\
  flow with a & & with value $F$ \\
  given value & & \\
  \hline
  Finding a      & An integer $F$, & An integer flow $f$ with value $F$, \\
  minimum-cost   & a scenario $s$ & whose cost (under scenario $s$) \\
  flow           & & is minimal \\
  \hline
\end{tabular}
\caption{Specification of our basic procedures with flows.}
\end{center}
\end{table}

In the remaining part of this section we briefly describe how
the procedures from Table~1 can be implemented. In many cases,
implementation is based on finding a path in a network. Depending
 on a particular procedure, here follow the details.
\begin{itemize}
  \item {\em Summation of flows\/} is implemented in a straightforward
        way. The same is true for {\em flow centering\/}.
  \item To implement {\em flow decomposition\/}, we consider the
        network consisting of arcs that are used by $f$, i.e.\
        arcs whose arc flow values according to $f$ are $>0$.
        Within that network we find a path from the source to the
        sink, e.g.\ with Moore's BFS algorithm \cite{Jung}. The
        found path determines the first unit flow $\phi_1$. We
        subtract $\phi_1$ from $f$, thus obtaining a simpler integer
        flow $f_1$ with value $F-1$. We repeat the same routine
        on $f_1$ in order to obtain the second unit flow $\phi_2$,
        \ldots , etc., until all $F$ unit flows are found.
  \item {\em Flow augmentation\/} is achieved in the following way. We
        construct the so-called {\em displacement\/} or {\em residual\/}
        network associated with $f$ \cite{Carre,Jung,Korte}. Such
        network consists of forward arcs (showing where $f$ can be
        increased) and of backward arcs (indicating where $f$ can
        be decreased). We try to find an arbitrary path in the
        displacement network from the source to the sink. For this
        purpose we use e.g.\ Moore's BFS or Tarjan's DFS algorithm
        \cite{Jung}. If the sought path does not exist, then $F$
        cannot be increased since it is already maximal. Otherwise,
        in order to obtain $g$, we modify $f$ according to the found
        path, so that the same integral amount of flow is added to
        each forward arc and subtracted from the inverse of each
        backward arc. The used amount of flow is chosen so that
        it does not exceed the limit imposed by arc capacities
        along the path.
  \item To implement {\em flow rounding\/}, we consider the pseudo-flow
        $\bar f$ obtained from $f$ by rounding all its arc values
        $x_{ij}$ to $\lfloor x_{ij}+0.5\rfloor$ (i.e.\ to the nearest
        integer). Note that $\bar f$ may violate the conservation
        rule in vertices, but it still must obey the capacity
        constraints along arcs due to the fact that all arc
        capacities are integral. We construct the network
        consisting of arcs whose arc flow values according to $\bar f$
        are $>0$. In that network we try to find a path from the source
        to the sink, by employing e.g.\ Moore's BFS or Tarjan's DFS
        algorithm \cite{Jung}. The found path determines the first unit
        flow $\phi_1$. We subtract $\phi_1$ from $\bar f$, thus obtaining
        a simpler pseudo-flow $\bar f_1$ with smaller value. We repeat
        the same routine on $\bar f_1$ in order to obtain the second
        unit flow $\phi_2$, \ldots, etc. Iteration stops after
        $\lfloor F+0.5\rfloor$ steps, or earlier if the remaining
        network becomes disconnected (i.e.\ if there is no path from
        the source to the sink). Then we sum up the collected unit
        flows $\phi_1, \phi_2, \ldots $ in order to obtain the first
        version of $g$. It is obvious that the obtained $g$ must be
        a feasible integer flow, namely it satisfies the flow conservation
        rule (since it is a sum of unit flows) and it also obeys the
        capacity constraints (since its arc values cannot exceed the
        corresponding values in $\bar f$). Finally, if the value of
        $g$ is less than $\lfloor F+0.5 \rfloor$, we repeatedly modify
        $g$ by using the previously described flow augmentation procedure
        until the desired flow value is reached.
  \item {\em Flow composition\/} can be done in the following way. A
        combined list of $\leq F$ unit flows is formed by selecting elements
        from the first and from the second input list in alternation
        (a randomly chosen element from the first list, then a randomly
        chosen element from the second list, then again from the first
        list, \ldots, etc.). During the whole construction process, the
        already selected unit flows must be compatible in the sense that
        their sum is a feasible flow. If in some step compatibility is
        violated, then that step may be repeated by randomly choosing
        some other element from the same input list. However, even
        with such repeated trials, it can still happen that the combined
        list cannot be completed to length $F$ because any remaining
        element from the appropriate input list would lead to
        incompatibility. In that case, construction stops with the
        combined list containing $<F$ unit flows. Anyway, the elements
        from the combined list are summed up in order to produce the
        first version of the output flow $h$. If the obtained $h$ has
        the value $<F$, it is repeatedly modified by the flow augmentation
        procedure until the desired value $F$ is reached.
  \item To achieve {\em flow cost reduction\/}, we try to find a
        negative-length cycle in the corresponding displacement network
        (considering unit costs as arc lengths, assuming that costs of
        forward arcs have positive signs and costs of backward arcs
        negative signs). To find such cycle we can use e.g.\ the
        Floyd-Warshall algorithm \cite{Carre,Jung,Korte}. The remaining
        details regarding how the found cycle is used to transform $f$
        into $g$ are similar as for flow augmentation.
  \item To realize {\em flow perturbation\/}, we try to find an arbitrary
        cycle in the corresponding displacement network - it can be done
        e.g.\ by the backtracking DFS algorithm \cite{Jung}. The remaining
        steps needed to obtain $g$ from $f$ are analogous as for flow cost
        reduction.
  \item {\em Flow harmonization\/} can be implemented as flow perturbation
        applied to $f$. However, a customized displacement network
        associated with $f$ is applied. Namely, only those forward
        arcs are taken into account that are used by $g$. Also, only
        those backward arcs are considered that correspond to arcs
        not used by $g$.
  \item To find a {\em flow with a given value\/}, we could start from the
        null flow $f$ and transform it into a flow with value $F$
        by repeated application of the flow augmentation procedure.
        There exist also more compact and more efficient implementations,
        such as the Edmonds-Karp, Dinic or Malholtra-Kumar-Maheshwari
        algorithm \cite{Jung,Korte}.
  \item In order to find a {\em minimum-cost flow\/}, we could start from
        an arbitrary flow $f$ with value $F$, e.g.\ the one obtained
        by the previously described procedure. Then we could transform
        $f$ into a minimum-cost flow by iterating the flow cost reduction
        procedure. Again, there exist more compact and more efficient
        implementations. Some of them are based on finding {\em shortest
        paths\/} in networks (considering arc unit costs as their lengths).
        Finding shortest paths can be implemented by Dijkstra's algorithm
        (if arc lengths are nonnegative) or by the Bellman-Ford algorithm
        (if arc lengths can be negative but there is still no cycle with
        negative length) \cite{Carre,Jung}.
\end{itemize}

%
%        Section 4
%

\section{Heuristic solutions}

In this section we first describe our {\em local search\/}
algorithm for solving the RMCIF problem. Its outline is specified
by the pseudo-code shown in Figure~2. Our pseudo-code follows the
well-known overall strategy described e.g.\ in \cite{Papa,Talbi}.
Thus for a given problem instance, the algorithm starts by finding
a feasible flow that will serve as the {\em initial solution\/},
i.e. the first version of the {\em current solution\/}. Then the
current solution is iteratively improved. In each iteration, the
so-called {\em neighborhood\/} of the current solution is generated.
The neighborhood consists of feasible flows obtained by modifying the
current flow in certain ways. All flows in the neighborhood are evaluated
according to the chosen robust criterion of optimality (objective function).
The best-evaluated member of the neighborhood is identified. If the
best-evaluated member is better then the current solution, it
becomes the new current solution and the algorithm proceeds with
another iteration. Otherwise the algorithm stops and proclaims the
last current solution to be nearly optimal (in the robust
sense) for the considered problem instance.

\begin{figure}[hbt]
\begin{center}\small
\begin{picture}(400,300) (0,0)
\put(0,0){\framebox(400,290)
{\tt \shortstack[l]{
Local search for the RMCIF problem \{ \\
\hspace*{10 pt} \\
\hspace*{10 pt} import the predefined parameters $neighborhoodSize$, $iterationLimit$; \\
\hspace*{10 pt} input the RMCIF problem instance; \\
\hspace*{10 pt} $CurrentFlow$ = an initial feasible flow; \\
\hspace*{10 pt} evaluate $CurrentFlow$ according to the \\
\hspace*{10 pt} robust criterion of optimality; \\
\hspace*{10 pt} $iterationCount$ = 0; \\
\hspace*{10 pt} \\
\hspace*{10 pt} while ($iterationCount$ < $iterationLimit$) \{ \\
\hspace*{10 pt} \\
\hspace*{30 pt} generate the neighborhood $\cal N$ of $CurrentFlow$ \\
\hspace*{30 pt} consisting of $neighborhoodSize$ feasible flows; \\
\hspace*{30 pt} evaluate all flows in $\cal N$ according to the \\
\hspace*{30 pt} robust criterion of optimality; \\
\hspace*{30 pt} $BestFlow$ = the best-evaluated flow in $\cal N$; \\
\hspace*{30 pt} if ($BestFlow$ is better than $CurrentFlow$) \\
\hspace*{50 pt} $CurrentFlow$ = $BestFlow$; \\
\hspace*{30 pt} else \\
\hspace*{50 pt} break; \\
\hspace*{30 pt} $iterationCount$ += 1; \\
\hspace*{10 pt} \\
\hspace*{10 pt} \} \\
\hspace*{10 pt}  \\
\hspace*{10 pt} output $CurrentFlow$ as the solution; \\
\}
}}}
\end{picture}
\caption{Generic pseudo-code of local search.}
\end{center}
\end{figure}

As indicated by Figure~2, our local-search algorithm relies on two
predefined parameters: the first of them determines the neighborhood
size, and the second is used within a stopping condition. We see
that the whole search can in fact terminate in two ways: (normally)
when the current flow cannot be improved any more, or (exceptionally)
when a predefined limit for total number iterations is reached.

According to the outline shown in Figure~2, there are many possible
variants of local search. First of all, the variants can differ in
the chosen robust criterion of optimality. Indeed, we can choose
absolute robustness according to the problem specification RMCIF-A,
or robust deviation captured by RMCIF-D. Note that in the case of
robust deviation the cost of the minimum-cost flow for each particular
scenario is needed within the objective function. Those costs should
be computed in advance by using the appropriate procedure from the
previous section.

After the robust criterion has been chosen, there is still a lot of
possibilities how to construct the initial feasible flow or generate
the neighborhood of the current flow. In this paper we restrict to
four local-search variants for each robustness criterion. The variants
(i.e.\ heuristics) are called LS-1, LS-2, LS-3 and LS-4, respectively.
Their properties are summarized in Table~2.

\begin{table}[hbt]
\begin{center}
\begin{tabular}{|l||l|l|l|}
  \hline
               & Initial                & Neighborhood   & Repeated       \\
               & flow                   & generation     & execution?     \\
  \hline\hline
  LS-1         & arbitrary              & iterated flow  & no             \\
               & feasible flow          & cost reduction &                \\
  \hline
  LS-2         & best-evaluated         & iterated flow  & no             \\
               & minimum-cost flow      & cost reduction &                \\
  \hline
  LS-3         & rounded centered       & iterated flow  & no             \\
               & minimum-cost flow      & cost reduction &                \\
  \hline
  LS-4         & minimum-cost flow for  & iterated flow  & yes - once for \\
               & a particular scenario  & cost reduction & each scenario  \\
  \hline
\end{tabular}
\caption{Variants of local search.}
\end{center}
\end{table}

As it can be seen from Table~2, our four heuristics based on local
search start from different initial flows. However, the neighborhood
of a current flow is always generated in the same manner. One of
the heuristics involves repeated execution of the pseudo-code
from Figure~2. All initial flows and neighborhoods are produced by
using appropriate basic procedures from the previous section. Here
are some more details.
\begin{itemize}
\item In LS-1 the initial flow is constructed in
      a rather {\em rudimentary\/} way. The idea is to choose
      an arbitrary feasible flow whose design is not influenced
      by the given scenarios. It can be obtained by finding
      a flow with a given value.
\item In LS-2 the initial flow is chosen by a kind of
      {\em greedy\/} approach. First, the minimum-cost flow
      for each scenario is found. Then, the obtained minimum-cost
      flows are evaluated according to the chosen robust
      criterion of optimality. Finally, the best-evaluated
      flow is selected.
\item In LS-3 the initial flow is constructed by
      using a {\em balanced\/} approach. Again, the minimum-cost
      flow for each scenario is found. Next, the centered flow
      based on all those minimum-cost flows is computed. Finally,
      flow rounding is applied to the centered flow in order to
      make it integral.
\item In LS-4 initialization is subject to {\em multiple\/}
      trials. The initial flow is again chosen as the minimum-cost
      flow for a particular scenario. However, the whole pseudo-code
      from Figure~2 is executed separately for each scenario.
      Finally, among all solutions obtained in this manner the
      best one (according to the robust criterion) is chosen as
      the final solution.
\item In all considered variants of local search, neighborhood
      generation is based on flow cost reduction according to
      {\em various\/} scenarios. Indeed, the current flow is
      cost-reduced separately according to each scenario. Thus
      the neighborhood initially consists of as many flows as
      there are scenarios. In order to obtain a larger neighborhood,
      cost reduction for a particular scenario can be iterated
      several times and all intermediate flows can be recorded.
\end{itemize}

In the remaining part of this section we describe our
{\em evolutionary\/} algorithm for solving the RMCIF problem. Its
outline is given by the pseudo-code in Figure~3. Our pseudo-code
follows the well-known overall strategy described e.g.\ in
\cite{Eiben,Talbi}. Indeed, it is a randomized computing
process which maintains a {\em population\/} of feasible solutions
(flows). The population is iteratively changed, thus producing
a series of population versions called {\em generations\/}. All
flows in a generation are evaluated according to the chosen robust
criterion of optimality (objective function). We expect that the
best-evaluated flow in the last generation should represent a nearly
optimal solution (in the robust sense) for the considered problem
instance.

\begin{figure}[p]
\begin{center}\small
\begin{picture}(420,550) (0,0)
\put(0,0){\framebox(420,550)
{\tt \shortstack[l]{
Evolutionary computing for the RMCIF problem \{ \\
\hspace*{10 pt} \\
\hspace*{10 pt} import the predefined parameters $populationSize$, $generationLimit$, \\
\hspace*{10 pt} $noImprovementLimit$, $similarityThreshold$, $mutationThreshold$; \\
\hspace*{10 pt} input the RMCIF problem instance; \\
\hspace*{10 pt} initialize the population $\cal P$ so that it \\
\hspace*{10 pt} consists of $populationSize$ feasible flows; \\
\hspace*{10 pt} evaluate all flows in $\cal P$ according to the \\
\hspace*{10 pt} robust criterion of optimality; \\
\hspace*{10 pt} $generationCount$ = 0; \\
\hspace*{10 pt} $noImprovementCount$ = 0; \\
\hspace*{10 pt} \\
\hspace*{10 pt} while ( ($generationCount$ < $generationLimit$) \&\& \\
\hspace*{55 pt} ($noImprovementCount$ < $noImprovementLimit$) ) \{ \\
\hspace*{10 pt} \\
\hspace*{30 pt} // crossover \\
\hspace*{30 pt} $Parent1$ = a "good" flow from $\cal P$ selected by tournament; \\
\hspace*{30 pt} $Parent2$ = another "good" flow from $\cal P$ selected by tournament; \\
\hspace*{30 pt} $Child$ = crossover of $Parent1$ and $Parent2$; \\
\hspace*{30 pt} leave $Parent1$ and $Parent2$ in $\cal P$; \\
\hspace*{30 pt} evaluate $Child$ according to the \\
\hspace*{30 pt} robust criterion of optimality; \\
\hspace*{30 pt} insert $Child$ into $\cal P$ by taking \\
\hspace*{30 pt} into account $similarityThreshold$; \\
\hspace*{10 pt} \\
\hspace*{30 pt} // mutation \\
\hspace*{30 pt} $rand$ = a random integer between 1 and 100; \\
\hspace*{30 pt} if ($rand$ <= $mutationThreshold$) \{ \\
\hspace*{50 pt} $Parent$ = a randomly chosen flow \\
\hspace*{50 pt} from $\cal P$ which is not the best in $\cal P$; \\
\hspace*{50 pt} $Mutant$ = mutation of $Parent$; \\
\hspace*{50 pt} evaluate $Mutant$ according to the \\
\hspace*{50 pt} robust criterion of optimality; \\
\hspace*{50 pt} replace $Parent$ in $\cal P$ with $Mutant$;\\
\hspace*{30 pt} \} \\
\hspace*{30 pt} $generationCount$ +=1; \\
\hspace*{30 pt} if (the best-evaluated flow in $\cal P$ is now better \\
\hspace*{52 pt} than it was in the previous generation) \\
\hspace*{50 pt} $noImprovementCount$ = 0; \\
\hspace*{30 pt} else \\
\hspace*{50 pt} $noImprovementCount$ += 1; \\
\hspace*{10 pt} \\
\hspace*{10 pt} \} \\
\hspace*{10 pt}  \\
\hspace*{10 pt} output the best-evaluated flow in $\cal P$ as the solution; \\
\}
}}}
\end{picture}
\caption{Generic pseudo-code of evolutionary computing.}
\end{center}
\end{figure}

The most important elements of the algorithm from Figure~3 are
the {\em evolutionary operators\/} that produce new flows from
old ones, thus changing the current population. There is a unary
operator, called {\em mutation\/}, which makes a small change
of a single flow. There is also a binary operator, called
{\em crossover\/}, which creates a new flow (child) by combining
parts of two existing flows (parents). Also an important element
of the algorithm is the {\em initialization procedure\/}, which
generates the initial population of flows.

Note that evolutionary computing described by Figure~3 relies
on five predefined parameters. The first of them determines
the population size, and the next two are used within stopping
conditions. We see that the whole computation can stop in
two ways: either when the total number of generations reaches
a predefined limit, or when the best solution does not improve
during a predefined number of consecutive generations. The last
but one parameter is used within the procedure for inserting
children into the population, as it will be explained later.
The last parameter controls intensity of using mutation.

An additional element of our evolutionary algorithm is selection
of a ``good" (or ``bad") flow from the population. It is accomplished
by so-called {\em tournament selection\/} \cite{Eiben,Talbi}. Indeed,
several flows are picked up randomly, and then the best-evaluated (or
the worst-evaluated) among them is selected.

Yet another part of our evolutionary computing is the
{\em insertion procedure\/} used to insert newly produced solutions
(children) into the current population, while keeping population
size constant. As it is suggested by the pseudo-code in Figure~3,
our insertion procedure relies on the concept of similarity and
uses the predefined parameter called {\em similarityThreshold\/}.
We say that two flows are {\em similar\/} if the difference of
their robust costs (robust objective function values), expressed
as a percentage of the best robust cost within the population,
is not greater than {\em similarityThreshold\/}. Insertion
``by taking into account {\em similarityThreshold\/}" means
the following. If there exists another flow in the population
that is similar in the above sense to the new one, then the
better of those two ``twins" is retained and the other one
discarded. If there is no similar flow, then the new one is
retained, and some other ``bad" flow from the population is
selected by tournament and discarded.

The outline shown in Figure~3 allows many possible variants, thus
producing slightly different evolutionary heuristics. First of all,
the variants can differ in the chosen robust criterion of optimality
- it can be absolute robustness according to RMCIF-A or robust
deviation specified by RMCIF-D. Again, in the case of robust
deviation the minimal flow costs for particular scenarios have
to be computed in advance.

After the robust criterion has been chosen, there are still
different options regarding how the population is initialized
or how the evolutionary operators are implemented. In this
paper we study nine evolution variants for each robustness
criterion. The variants (heuristics) are called EC-1, EC-2,
\ldots , EC-9, respectively. Their properties are presented
in Table~3.

\begin{table}[hbt]
\begin{center}
\begin{tabular}{|l||l|l|l|}
  \hline
               & Initial               & Crossover           & Mutation     \\
               & population            & operator            & operator     \\
  \hline\hline
  EC-1         & mutated               & flow centering,     & flow         \\
               & minimum-cost flows    & flow rounding       & perturbation \\
  \hline
  EC-2         & mutated               & flow centering,     & flow cost    \\
               & minimum-cost flows    & flow rounding       & reduction    \\
  \hline
  EC-3         & mutated               & flow centering,     & local        \\
               & minimum-cost flows    & flow rounding       & search       \\
  \hline
  EC-4         & mutated               & flow                & flow         \\
               & minimum-cost flows    & harmonization       & perturbation \\
  \hline
  EC-5         & mutated               & flow                & flow cost    \\
               & minimum-cost flows    & harmonization       & reduction    \\
  \hline
  EC-6         & mutated               & flow                & local        \\
               & minimum-cost flows    & harmonization       & search       \\
  \hline
  EC-7         & mutated               & flow decomposition, & flow         \\
               & minimum-cost flows    & flow composition    & perturbation \\
  \hline
  EC-8         & mutated               & flow decomposition, & flow cost    \\
               & minimum-cost flows    & flow composition    & reduction    \\
  \hline
  EC-9         & mutated               & flow decomposition, & local        \\
               & minimum-cost flows    & flow composition    & search       \\
  \hline
\end{tabular}
\caption{Variants of evolutionary computing.}
\end{center}
\end{table}

As shown in Table~3, our nine heuristics based on evolutionary
computing use essentially the same method to initialize the population.
However, they employ three different crossover operators and three
mutation operators. All initializations, crossovers and mutations are
realized by combining basic procedures from the previous section. Here
are some details.
\begin{itemize}
  \item In EC-1, EC-2 and EC-3 the crossover of two flows $f$
        and $g$ is obtained in a relatively {\em straightforward\/}
        way. First, the centered version of $f$ and $g$ is
        computed. Then, flow rounding of the centered flow
        is performed in order to achieve integrality.
        The rounded centered flow is output as the child.
  \item In EC-4, EC-5 and EC-6 the crossover of two flows
        $f$ and $g$ is obtained in a {\em more subtle\/} way.
        Namely, the child is produced by flow harmonization
        applied to $f$ and $g$.
  \item In EC-7, EC-8 and EC-9 the crossover of two flows $f$
        and $g$ is constructed in a rather {\em complex\/} way.
        The construction starts with flow decomposition, which
        breaks both $f$ and $g$ into lists of unit flows. Those
        two lists are then recombined through flow composition
        in order to produce the child flow.
  \item In EC-1, EC-4 and EC-7 the mutant of a flow $f$ is
        obtained by applying flow perturbation to $f$.
  \item In EC-2, EC-5 and EC-8 the mutant of a flow $f$ is
        obtained by applying flow cost reduction to $f$
        according to a randomly chosen scenario $s$.
  \item In EC-3, EC-6 and EC-9 mutation is implemented as
        a full-scale local search. Thus the mutant of a flow
        $f$ is produced by running the whole local search
        algorithm from Figure~2 with $f$ taken as the initial
        solution. Consequently, EC-3, EC-6 and EC-9 can be
        regarded as hybrids of evolution and local search.
  \item In all evolution variants, the initial population is
        generated in the following way. First, the minimum-cost
        flow for each scenario is constructed. All obtained flows,
        being optimal for particular scenarios, are inserted into
        the population. If such population happens to be too small,
        additional members are produced from the original ones
        by applying mutation. Thereby the same mutation operator
        is used as in the main part of the algorithm. Even more
        population members can be produced by successive mutation,
        or by collecting intermediate solutions if mutation is
        implemented by local search.
\end{itemize}

%
%        Section 5
%

\section{Experimental results}

In order to perform experiments, we have implemented our four variants
of local search and nine variants of evolutionary computing (altogether
thirteen heuristics) as a single C\# program \cite{Troelsen}. Each particular
heuristic has been assembled from basic procedures according to Table~2
or 3 from Section~4. The basic procedures themselves have been implemented
as suggested in Section~3. Thereby the following standard networking
algorithms \cite{Jung, Korte} have been employed:
\begin{itemize}
  \item the Malholtra-Kumar-Maheshvari algorithm for finding flows with
        maximal or given values,
  \item Moore's BFS algorithm for finding paths with minimum number of arcs,
  \item Dijskstra's and the Bellman-Ford algorithm for finding shortest paths,
  \item the Floyd-Warshall algorithm for finding negative-length cycles,
  \item the backtracking DFS algorithm for finding arbitrary cycles.
\end{itemize}

The implemented high-level algorithms for local search and evolutionary
computing follow the generic pseudo-codes from Figures~2 and 3,
respectively. All parameters from the pseudo-codes are supported.
The program is always configured with certain fixed values
of those parameters, but at any time it can easily be reconfigured
with some other values.

The configured program takes as input the specification of a
concrete RMCIF problem instance, i.e.\
\begin{itemize}
  \item the set of vertices $V$ and the set of arcs $A$ with
        arc capacities $u_{ij}$,
  \item the desired flow value $F$,
  \item the set of scenarios $S$ with arc unit costs $c_{ij}^s$
        for each scenario,
  \item the minimal flow costs $z^s$ for particular scenarios
        computed in advance.
\end{itemize}
The given RMCIF problem instance is solved repeatedly:
\begin{itemize}
  \item according to both problem variants, RMCIF-A and
        RMCIF-D, respectively,
  \item with all thirteen heuristics, i.e.\ LS-1, LS-2, \ldots,
  LS-4, EC-1, EC-2, \ldots, EC-9.
\end{itemize}
For each combination of problem variant and heuristic the
program produces the following output:
\begin{itemize}
  \item full specification of the solution, i.e.\ the
        list of arc flow values $x_{ij}$ and the value
        of the robust objective function,
  \item time in seconds taken by the program to produce
        the solution.
\end{itemize}

Our program has been installed on a standard notebook computer with
a 2.60 GHz Intel Core i5-6440HQ processor and 4 GBytes of memory.
After some preliminary testing, the program parameters have been
set in the following way: {\em neighborhoodSize\/} = 30,
{\em iterationLimit\/} = $\infty$, {\em populationSize\/} = 30,
{\em generationLimit\/} = $\infty$, {\em noImprovementLimit\/} = 300,
{\em similarityThreshold\/} = 5, {\em mutationThreshold\/} = 1.

In order to measure accuracy of approximate solutions obtained by
heuristics, we have also implemented an exact solver for the RMCIF
problem. It is based on the well-known general-purpose optimization
package IBM ILOG CPLEX \cite{Ibm}. Thanks to that additional software,
it was possible to obtain exact solutions (i.e.\ solutions that are
truly optimal in the robust sense) for smaller problem instances.
Thereby the ``linearized" problem versions RMCIF-A$^\prime$ and
respectively RMCIF-D$^\prime$ were solved. The exact solver takes
similar input data as our heuristics, produces as output the exact
robust solution costs for both robustness criteria, and measures
its own computing time during each computation. It has been installed
on the same hardware as the C\# program that runs heuristics.

In our experiments, we have used a set of 30 carefully constructed
RMCIF problem instances, whose identifiers are I-01, I-02, \ldots,
I-30, respectively. They have been chosen as large enough to
be nontrivial, but still small enough to be solvable by the exact
solver. It means that slightly larger instances (more vertices,
arcs or scenarios) cannot anymore be solved to optimality, at least
not with our hardware and software. Namely, on such larger instances
the exact solver fails due to memory overflow.

All our problem instances involve layered networks similar to those
occurring in standard applications of network flows. The idea is
illustrated by Figure~4. Thus all vertices except the source and
the sink are distributed among layers. Arcs can connect only
vertices between adjacent layers. Layer width (number of vertices
within a layer) can be fixed (the same for each layer) or varying.

\begin{figure}[hbt]
\begin{center}
    \ \\
    \ \\
    \leavevmode
    \includegraphics[width=420pt]{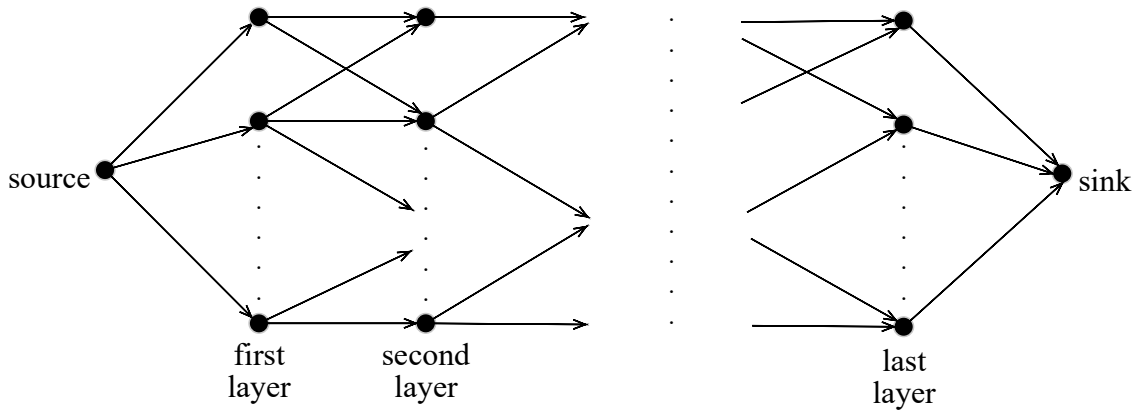}
    \caption{Structure of a layered network.}
\end{center}
\end{figure}

Important properties of our problem instances are summarized in
Table~4. The values shown in the table have been chosen by hand,
and the remaining details not visible from the table (such as
actual configuration of arcs or actual values of $u_{ij}$ or
$c_{ij}^s$) have been generated randomly. The complete specification
of each problem instance can be obtained upon a request from the
first author by e-mail.

\begin{table}[hbt]
\begin{center}
{\normalsize
\setlength{\tabcolsep}{1ex}
\begin{tabular}{|c|c|c|c|c|c|c|c|}
\hline
 Instance   & Number of  & Number  & Number of & Flow  & Number & Layer & Range             \\
 identifier & vertices   & of arcs & scenarios & value & of     & width & for               \\
            & $|V|$      & $|A|$   & $|S|$     & $F$   & layers &       & $u_{ij},c_{ij}^s$ \\
\hline\hline
I-01  & 18 & 80  & 30 & 243 & 2  & 8 (fixed) & $0-99$ \\
\hline
I-02  & 18 & 56  & 30 & 177 & 4  & 4 (fixed) & $0-99$ \\
\hline
I-03  & 18 & 32  & 30 & 93  & 8  & 2 (fixed) & $0-99$ \\
\hline
I-04  & 18 & 34  & 30 & 20  & 7  & 3,2,2,2,2,2,3 & $0-99$ \\
\hline
I-05  & 17 & 60  & 30 & 177 & 3  & 5 (fixed) & $0-99$ \\
\hline
I-06  & 17 & 42  & 30 & 134 & 5  & 3 (fixed) & $0-99$ \\
\hline
I-07  & 17 & 59  & 30 & 198 & 2  & 11,4 & $0-99$ \\
\hline
I-08  & 24 & 143 & 15 & 456 & 2  & 11 (fixed) & $0-99$ \\
\hline
I-09  & 24 & 44  & 15 & 48  & 11 & 2 (fixed) & $0-99$ \\
\hline
I-10 & 24 & 130 & 15 & 104 & 3  & 10,9,3 & $0-99$ \\
\hline
I-11 & 23 & 112 & 15 & 270 & 3  & 7 (fixed) & $0-99$ \\
\hline
I-12 & 23 & 60  & 15 & 52  & 7  & 3 (fixed) & $0-99$ \\
\hline
I-13 & 23 & 72  & 15 & 34  & 3  & 7,3,11 & $0-99$ \\
\hline
I-14 & 27 & 110 & 10 & 216 & 5  & 5 (fixed) & $0-99$ \\
\hline
I-15 & 27 & 153 & 10 & 78  & 3  & 3,8,14 & $0-99$ \\
\hline
I-16 & 26 & 168 & 10 & 506 & 2  & 12 (fixed) & $0-99$ \\
\hline
I-17 & 26 & 144 & 10 & 270 & 3  & 8 (fixed) & $0-99$ \\
\hline
I-18 & 26 & 120 & 10 & 228 & 4  & 6 (fixed) & $0-99$ \\
\hline
I-19 & 26 & 88  & 10 & 229 & 6  & 4 (fixed) & $0-99$ \\
\hline
I-20 & 26 & 69  & 10 & 112 & 8  & 3 (fixed) & $0-99$ \\
\hline
I-21 & 26 & 48  & 10 & 108 & 12 & 2 (fixed) & $0-99$ \\
\hline
I-22 & 26 & 118 & 10 & 16  & 5  & 2,4,6,8,4 & $0-99$ \\
\hline
I-23 & 30 & 224 & 5  & 628 & 2  & 14 (fixed) & $0-99$ \\
\hline
I-24 & 30 & 161 & 5  & 247 & 4  & 7 (fixed) & $0-99$ \\
\hline
I-25 & 30 & 104 & 5  & 218 & 7  & 4 (fixed) & $0-99$ \\
\hline
I-26 & 30 & 56  & 5  & 63  & 14 & 2 (fixed) & $0-99$ \\
\hline
I-27 & 30 & 198 & 5  & 97  & 3  & 7,10,11 & $0-99$ \\
\hline
I-28 & 29 & 180 & 5  & 356 & 3  & 9 (fixed) & $0-99$ \\
\hline
I-29 & 29 & 78  & 5  & 97  & 9  & 3 (fixed) & $0-99$ \\
\hline
I-30 & 29 & 90  & 5  & 30  & 7  & 2,5,3,4,6,2,5 & $0-99$ \\
\hline
\end{tabular}
}
\caption{Properties of the chosen RMCIF problem instances.}
\end{center}
\end{table}

Regarding our choice of problem instances, one may argue that it would
be much better if we used some standard benchmark data instead of our
own synthetic data. This is certainly true. However, the obstacle is
that, to the best of our knowledge, a suitable benchmark collection
does not exist. There are some well-known repositories of minimum-cost
flow problem instances available on Internet, e.g.\ \cite{Beasley,Dimacs},
but they deal only with the conventional (non-robust) problem variant.
Expanding a conventional benchmark instance with more scenarios would
not make sense since it would produce a completely new instance that
cannot be regarded as a benchmark any more.

In our experiments we have run our C\# program on all problem instances
from Table~4. Consequently, each problem instance has been solved
according to each of the two robust optimality criteria by each of
the thirteen heuristics. The obtained solutions have been stored
in a data file, which is too voluminous to be reproduced here but
is again available from the first author by e-mail. Still, some
excerpts from that file are visible in Table~5. Indeed, for each
problem instance and each robustness criterion the table presents
a selected approximate solution. It is the one with the smallest
robust cost among those whose computing time is at most half of
the time needed by the exact solver. To allow comparison, the
table also contains the corresponding exact solutions. Any
solution, either exact or approximate, is described by its
robust cost (a whole number) and its computing time in seconds
(a decimal number). In case of an approximate solution, the
involved heuristic is also identified.

\begin{table}[hbt]
\begin{center}
{\normalsize
\setlength{\tabcolsep}{0.7ex}
\begin{tabular}{|c|cc|ccc|cc|ccc|}
\hline
 Instance & \multicolumn{2}{c|}{RMCIF-A} & \multicolumn{3}{c|}{RMCIF-A} &
 \multicolumn{2}{c|}{RMCIF-D} & \multicolumn{3}{c|}{RMCIF-D} \\
 identifier & \multicolumn{2}{c|}{exact} & \multicolumn{3}{c|}{approximate} &
 \multicolumn{2}{c|}{exact} & \multicolumn{3}{c|}{approximate} \\
            & \multicolumn{2}{c|}{solution} & \multicolumn{3}{c|}{solution} &
 \multicolumn{2}{c|}{solution} & \multicolumn{3}{c|}{solution} \\
\hline\hline
I-01 & 39700 & 5.86s & EC-9 & 40510 & 2.79s & 10612 & 2557.40s & EC-7 & 11115 & 3.37s \\
\hline
I-02 & 47676 & 29.23s & EC-8 & 48228 & 1.25s & 12746 & 404.24s & EC-9 & 13507 & 1.99s \\
\hline
I-03 & 52501 & 0.20s & EC-6 & 52642 & 0.06s & 5099 & 0.29s & EC-2 & 5291 & 0.06s \\
\hline
I-04 & 9473 & 0.25s & EC-8 & 9487 & 0.11s & 801 & 0.45s & EC-3 & 835 & 0.11s \\
\hline
I-05 & 37864 & 62.78s & EC-9 & 38420 & 1.79s & 10308 & 1545.33s & EC-7 & 10771 & 1.29s \\
\hline
I-06 & 43074 & 0.56s & EC-6 & 44823 & 0.16s & 10096 & 26.88s & EC-8 & 10580 & 1.86s \\
\hline
I-07 & 32636 & 5.06s & EC-9 & 32938 & 2.25s & 10641 & 48.87s & EC-9 & 11185 & 2.40s \\
\hline
I-08 & 68343 & 0.50s & EC-4 & 70835 & 0.15s & 16302 & 320.35s & EC-7 & 17340 & 18.97s \\
\hline
I-09 & 30351 & 0.23s & EC-1 & 31223 & 0.05s & 5871 & 0.39s & EC-8 & 6121 & 0.13s \\
\hline
I-10 & 18790 & 9.29s & EC-8 & 19750 & 1.77s & 8379 & 217.35s & EC-7 & 9226 & 2.98s \\
\hline
I-11 & 49736 & 101.12s & EC-9 & 52057 & 9.45s & 14987 & 485.17s & EC-9 & 16155 & 8.32s \\
\hline
I-12 & 19576 & 2.37s & EC-8 & 20327 & 0.47s & 8850 & 1.15s & EC-7 & 9761 & 0.51s \\
\hline
I-13 & 7852 & 0.25s & EC-1 & 8160 & 0.07s & 876 & 0.67s & EC-8 & 974 & 0.32s \\
\hline
I-14 & 59609 & 22.40s & EC-8 & 61452 & 6.31s & 18005 & 68.27s & EC-8 & 19671 & 6.77s \\
\hline
I-15 & 15388 & 0.92s & LS-2 & 16199 & 0.04s & 6177 & 9.97s & EC-9 & 6317 & 3.48s \\
\hline
I-16 & 81323 & 0.25s & EC-4 & 85928 & 0.07s & 15034 & 77.70s & EC-8 & 16328 & 29.02s \\
\hline
I-17 & 49328 & 61.03s & EC-8 & 52194 & 8.95s & 16579 & 800.00s & EC-9 & 18626 & 10.37s \\
\hline
I-18 & 53075 & 10.06s & EC-7 & 54322 & 4.19s & 16128 & 729.75s & EC-9 & 17166 & 9.17s \\
\hline
I-19 & 69116 & 5.89s & EC-8 & 72031 & 2.82s & 20122 & 18.12s & EC-7 & 22301 & 6.95s \\
\hline
I-20 & 48256 & 0.31s & EC-4 & 52440 & 0.03s & 12369 & 4.89s & EC-7 & 13765 & 1.29s \\
\hline
I-21 & 69159 & 0.25s & EC-5 & 70111 & 0.04s & 9120 & 0.32s & EC-2 & 10499 & 0.08s \\
\hline
I-22 & 4175 & 1.91s & EC-7 & 4347 & 0.16s & 2039 & 1.07s & EC-9 & 2124 & 0.50s \\
\hline
I-23 & 75982 & 0.35s & EC-5 & 80466 & 0.16s & 12821 & 6.25s & EC-6 & 17979 & 0.75s \\
\hline
I-24 & 49509 & 1.41s & EC-4 & 54611 & 0.12s & 16227 & 2.03s & EC-5 & 22624 & 0.09s \\
\hline
I-25 & 76696 & 0.29s & EC-5 & 85025 & 0.07s & 15115 & 3.06s & EC-5 & 20752 & 0.16s \\
\hline
I-26 & 46485 & 0.34s & EC-4 & 47452 & 0.02s & 5972 & 0.29s & LS-2 & 7211 & 0.02s \\
\hline
I-27 & 17451 & 1.17s & EC-5 & 18668 & 0.08s & 4191 & 1.34s & EC-5 & 5216 & 0.17s \\
\hline
I-28 & 66375 & 0.71s & LS-3 & 72481 & 0.03s & 13037 & 1.17s & EC-5 & 17504 & 0.11s \\
\hline
I-29 & 44382 & 0.57s & EC-4 & 46395 & 0.04s & 9203 & 0.42s & EC-5 & 11250 & 0.06s \\
\hline
I-30 & 11269 & 0.58s & EC-5 & 11737 & 0.04s & 2729 & 0.43s & EC-7 & 3302 & 0.20s \\
\hline
\end{tabular}
}
\caption{Solutions for the used RMCIF problem instances.}
\end{center}
\end{table}

The raw measurements from the described experiments could be processed
in many different ways. In the remaining part of this section we
present a simple analysis based on the whole collection of values
from our data file. Within our analysis we have computed the relative
error of each approximate solution, as the difference between the
approximate robust cost and the corresponding exact cost divided by the
exact cost. Also, we have computed the speedup of each approximate computation
as the quotient of the corresponding ``exact" computing time vs.\ the
``approximate" computing time. The obtained relative errors for a given
robust problem variant and a given heuristic have been averaged over
the whole set of problem instances. The same kind of averaging has also
been done with the speedups. All averages obtained in this way are
presented in Table~6. The table reveals ``typical" behavior of each
heuristic applied to each problem variant.

Now follows a discussion about the obtained results. The first thing we
can immediately observe is a  big difference between the RMCIF-A and the
RMCIF-D problem variant. Although the formulations of both variants look
similar, they behave quite differently. From Table~5 it is visible that
``exact" computing times for RMCIF-D are usually much larger than for
RMCIF-A, which means that RMCIF-D is harder to solve exactly. From
Table~6 it is visible that relative errors for RMCIF-D are about 4 times
larger than for RMCIF-A, which means that RMCIF-D is also harder to
approximate. Larger relative errors can partially be explained by the
following fact: the objective function in RMCIF-D measures deviations
from conventional costs, not actual costs, and takes therefore smaller
values than the objective function in RMCIF-A. Consequently, a certain
difference from the exact solution within RMCIF-D will produce a larger
relative error than it would produce within RMCIF-A.

\clearpage

\begin{table}[hbt]
\begin{center}
{\normalsize
\begin{tabular}{|l||cc|cc|}
\hline
 & \multicolumn{2}{c|}{RMCIF-A } & \multicolumn{2}{c|}{RMCIF-D} \\
\hline\hline
LS-1 & 14.40\% & 362.02$\times$ & 56.89\% & 25025.52$\times$ \\
\hline
LS-2 & 9.47\% & 500.43$\times$ & 41.04\% & 13789.07$\times$ \\
\hline
LS-3 & 10.80\% & 232.98$\times$ & 34.14\% & 10122.76$\times$ \\
\hline
LS-4 & 8.01\% & 26.27$\times$ & 35.62\% & 489.01$\times$ \\
\hline
EC-1 & 8.98\% & 52.94$\times$ & 34.48\% & 1389.52$\times$ \\
\hline
EC-2 & 8.89\% & 43.89$\times$ & 32.83\% & 975.39$\times$ \\
\hline
EC-3 & 7.49\% & 17.54$\times$ & 29.44\% & 584.88$\times$ \\
\hline
EC-4 & 7.13\% & 169.01$\times$ & 35.86\% & 4901.75$\times$ \\
\hline
EC-5 & 7.31\% & 151.64$\times$ & 28.55\% & 5013.30$\times$ \\
\hline
EC-6 & 6.35\% & 40.90$\times$ & 29.26\% & 1409.51$\times$ \\
\hline
EC-7 & 3.85\% & 4.81$\times$ & 15.17\% & 93.53$\times$ \\
\hline
EC-8 & 3.64\% & 4.60$\times$ & 14.49\% & 107.70$\times$ \\
\hline
EC-9 & 3.77\% & 3.14$\times$ & 13.18\% & 59.02$\times$ \\
\hline
\end{tabular}
}
\caption{Average errors (\%) and average speedups ($\times$) obtained
  for different combinations of heuristics vs.\ problem variants.}
\end{center}
\end{table}

The next important thing we can observe from our results is that local
search is in general faster but less accurate than evolutionary computing.
This is just what anybody would expect. However, there are some fine
differences in behavior of both types of heuristics regarding the two
problem variants.

Let us restrict for a moment to the part of Table~6 dealing with RMCIF-A.
Then we can see that local search applied to that problem variant can
obtain moderate precision (errors about 8\%), while evolutionary computing
can reach higher precision (errors about 4\%). Among different variants of
local search, the most accurate is LS-4. This is not a surprise since LS-4
is the only one with repeated execution, i.e.\ it repeats the whole searching
process several times with different initial flows. On the other hand, LS-4
is for the same reason the slowest among local-search variants, although it
still achieves a significant speedup (about 26$\times$) compared to exact
solving. Among different variants of evolutionary computing, the most
accurate are EC-7, EC-8 and EC-9. This is also not a surprise since EC-7,
EC-8 and EC-9 are again much more elaborate compared to the remaining
evolutionary variants. Namely, their crossover operator is based on full
decomposition of given flows into unit flows and composition of mutually
compatible unit flows into a new flow, which is time-consuming. Moreover, the
mutation operator in EC-9 is a  full-scale local search, which is also
relatively time-consuming. As a consequence, EC-7, EC-8 and EC-9 turn out
to be rather slow, i.e.\ they are in average only 3 to 5 times faster
than the exact solver. If we are looking for a tradeoff among accuracy
and speed, we should use a similar evolutionary variant EC-6, whose
accuracy is not much lower (errors about 6\%), but whose speedup is
higher (over 40$\times$).

Let us finally restrict to the part of Table~6 dealing with RMCIF-D.
Here, the relative errors are worse than for RMCIF-A, but the speedups
are better. Relative ranking of particular heuristics remains roughly
the same as for RMCIF-A. A small difference in ranking of local-search
variants is that LS-3 now seems to be slightly more accurate than LS-4.
Also, among the tree previously established most accurate evolutionary
variants, EC-9 now appears to be slightly better than the other two.
Precision of local search must be regarded as unsatisfactory (errors
greater than 34\%). Precision of the best evolutionary variant EC-9
is better, but not spectacular (errors about 13\%). A good news is
that EC-9, although being quite slow, now achieves a good speedup
(about 60$\times$) with respect to the exact solver.

%
%        Section 6
%

\section{Conclusions}

In this paper we have considered two robust variants of the minimum-cost
integer flow problem, i.e\ the absolute robust (min-max) and the robust
deviation (min-max regret) variant, respectively. Uncertainty in problem
formulation has been restricted to arc unit costs, and expressed by explicitly
given scenarios. Both problem variants turn out to be NP-hard, which justifies
their approximate solving.

As approximate solutions to the considered problem variants, thirteen heuristics
have been proposed, four of them based on local search, and nine on evolutionary
computing. All heuristics have experimentally been evaluated on a set of problem
instances big enough to be nontrivial but still small enough to be solved
exactly.

According to the obtained experimental results, there is a significant difference
among the two considered problem variants. Although both of them look similar on
the first sight, the second one is much harder to solve exactly, and even harder
to approximate. Therefore the results obtained for the absolute robust variant
are more satisfactory than those obtained for the robust deviation variant.

From the obtained experimental results we can also see that the heuristics based
on local search are fast but not very accurate. Luckily enough, better precision
is assured by evolutionary computing. Among nine evolutionary heuristics, the
best results are achieved by the one whose crossover operator is based on flow
decomposition/composition and whose mutation operator is in fact improvement by
local search. Such heuristic can as well be regarded as a hybrid of evolutionary
computing and local search.

A drawback of our most-precise heuristic is its slowness. Indeed, for
smaller absolute robust problem instances it is not much faster than an
exact algorithm. Still, its speedup improves and becomes satisfactory for
larger instances. Consequently, we believe that our most-accurate heuristic
will show its full potential on very large instances, i.e.\ in situations
where exact algorithms (or even their relaxed counterparts) fail.

In our future research, we plan to evaluate our heuristics on much larger
problem instances. Also, our plan is to extend our solutions to more general
minimum-cost integer flow problem variants, which would allow uncertainty
in arc capacities as well as in arc unit costs. A third direction of further
research would be to consider solving the same problems with other meta-heuristics,
such as simulated annealing or particle swarm optimization.

%
%        Acknowledgement
%

\section*{Acknowledgement}

This work has been fully supported by Croatian Science Foundation under the project IP-2018-01-5591.

%
%        References
%

\end{document}